\documentstyle[12pt]{article}
\catcode`@=11
\def\@bindentby#1{%
\begingroup\setbox3=\hbox{#1}\ifhmode\par\fi\noindent%
\dimen2=\linewidth\dimen4=\@totalleftmargin%
\copy3\unskip\advance\linewidth-\wd3\advance\@totalleftmargin\wd3%
\parshape=2\dimen4\dimen2\@totalleftmargin\linewidth%
\everypar{\parshape=1\@totalleftmargin\linewidth}%
\bgroup\ignorespaces}

\def\@eindentby{%
\egroup\advance\linewidth\wd3%
\advance\@totalleftmargin-\wd3%
\everypar{\parshape=1\@totalleftmargin\linewidth}%
\par\endgroup}

\def\fixindent{%
\egroup\everypar{\parshape=1\@totalleftmargin\linewidth}\par\bgroup%
}

\catcode`@=12

\catcode`@=11
\def\@tabsize{20pt}

\newcount\@lkind
\newcount\@icount
\newcount\@lcount
\newcount\@inmerge
\newcount\@algcount

\newdimen\@tabmargin
\newcounter{@glcount}

\newif\if@algmerged

\global\@lkind	   = 0
\global\@algcount  = 0
\global\@tabmargin = 0pt

\def\@noalgorithm{
	\bgroup\def\@alglabel{}\def\@algcurlabel{}\def\@algsep{}\def\@algprefix{}%
	\@lkind=0\@algmergedfalse\medskip\par
}

\def\@enoalgo{
	\egroup\everypar{\parshape=1\@totalleftmargin\linewidth\noindent}
}

\def\@enoalgorithm{
	\@enoalgo\medskip
}

\def\@balgorithm#1#2{

\par\global\advance\@algcount1\edef\@currentlabel{\number\@algcount}\setbox5=\hbox{\rm #1}
	\medskip\noindent{{\bf Algorithm} \number\@algcount :
\copy5(\hbox{#2})}\@noalgorithm
}

\def\@ealgorithm{\@enoalgo\vskip1.5pt\par\noindent{\bf End \box5}\medskip}

\def\@bsubroutine#1#2{
	\par\setbox5=\hbox{\rm #1}\medskip\noindent{{\bf Subroutine} :
\copy5(\hbox{#2})}
	\bgroup\def\@alglabel{}\def\@algcurlabel{}\def\@algsep{}\def\@algprefix{}
	\@lkind=0\@algmergedfalse\vskip1.5pt\par
}

\def\@bindentmerged{

\begingroup\@icount=0\let\item=\@iitem\let\noitem=\@nitem\let\mitem=\@mitem\@inmerge=1

\edef\@algprefix{\@algcurlabel}\@lcount=\value{@glcount}\setcounter{@glcount}{0}
}

\def\@bindent{

\begingroup\@icount=0\let\item=\@iitem\let\noitem=\@nitem\let\mitem=\@mitem\@inmerge=0

\advance\@tabmargin\@tabsize\advance\linewidth-\@tabsize\advance\@totalleftmargin\@tabsize
	\everypar{\parshape=1\@totalleftmargin\linewidth\noindent}
}

\def\@eindent{
	\ifnum0<\@icount\@eindentby\fi\ifnum\@inmerge=0
	\ifdim\@tabmargin>0pt\advance\@tabmargin-\@tabsize\fi
	\advance\linewidth\@tabsize\advance\@totalleftmargin-\@tabsize
	\everypar{\parshape=1\@totalleftmargin\linewidth\noindent}
	\fi\ifnum\@lcount>-1\setcounter{@glcount}{\@lcount}\fi\endgroup
}

\def\@bnonblank{

\@lcount=\value{@glcount}\setcounter{@glcount}{0}\@tabmargin=0pt\edef\@algprefix{}

}

\def\@listfill#1{\ifdim\@tabmargin>0pt\hbox{\kern-1.5em\kern-\@tabmargin\hbox
to1.5em{#1\hfil}%
	\unskip\kern\@tabmargin}\else\hbox to1.5em{#1\hfil}\fi
}

\def\@listbl{
	\@bindent\@lcount=-1\@algmergedfalse
}

\def\@listar{

\ifnum\@lkind=1\def\@algsep{.}\fi\edef\@alglabel{\@alglabel\@algcurlabel\@algsep}
	\@lkind=1\def\@algsep{}\if@algmerged\@bindentmerged
	\else
	\@bindent\@bnonblank\fi
}

\def\@listlc{
	\edef\@alglabel{\@alglabel\@algcurlabel\@algsep}\@lkind=2\def\@algsep{}
	\if@algmerged\@bindentmerged
	\else\@bindent\@bnonblank\fi
}

\def\@listuc{
	\edef\@alglabel{\@alglabel\@algcurlabel\@algsep}\@lkind=3\def\@algsep{}
	\if@algmerged\@bindentmerged
	\else\@bindent\@bnonblank\fi
}

\def\@arlabel{

\def\@algcurlabel{\arabic{@glcount}}\def\@algusedlabel{\@algprefix\@algcurlabel}
	\xdef\@currentlabel{\@alglabel\@algsep\@algcurlabel}
}

\def\@lclabel{
	\def\@algcurlabel{\alph{@glcount}}\def\@algusedlabel{\@algprefix\@algcurlabel}
	\xdef\@currentlabel{\@alglabel\@algsep\@algcurlabel}
}

\def\@uclabel{
	\def\@algcurlabel{\Alph{@glcount}}\def\@algusedlabel{\@algprefix\@algcurlabel}
	\xdef\@currentlabel{\@alglabel\@algsep\@algcurlabel}
}

\def\@aritem{
	\@arlabel\@bindentby{\@listfill{\@algusedlabel.}}
}

\def\@lcitem{
	\@lclabel\@bindentby{\@listfill{\@algusedlabel.}}
}

\def\@ucitem{
	\@uclabel\@bindentby{\@listfill{\@algusedlabel.}}
}

\def\@blitem{
	\@bindentby{\hbox to 0em{\hfil}}
}

\def\@iitem{
	\if@algmerged\@inmerge=0\@algmergedfalse\unskip\else
	\ifnum0<\@icount\unskip\@eindentby\fi\fi
	\advance\@icount1\global\stepcounter{@glcount}
	\ifcase\@lkind\@blitem\or\@aritem\or\@lcitem\or\@ucitem\fi\ignorespaces\unskip
}

\def\@nitem{

\ifnum0<\@icount\unskip\@eindentby\fi\advance\@icount1\@bindentby{\@listfill{\hbox{}}}%
	\ignorespaces\unskip
}

\def\@mitem{

\ifnum0<\@icount\unskip\@eindentby\fi\advance\@icount1\global\stepcounter{@glcount}
	\ifcase\@lkind\@bllabel\or\@arlabel\or\@lclabel\or\@uclabel\fi
	\@algmergedtrue\ignorespaces\unskip
}

\newenvironment{algorithm}[1]{\@balgorithm{#1}}{\@ealgorithm}

\newenvironment{listbl}{\@listbl}{\@eindent}

\catcode`@=12

\begin{document}
\title{An Extended Clustering Algorithm for Statistical Language Models}
\author{Joerg P. Ueberla\\
Forum Technology - DRA Malvern\\
St. Andrews Road, Malvern\\
Worcestershire, WR14 3PS, UK\\
email: ueberla@signal.dra.hmg.gb\\
DRA/CIS(CSE1)/RN94/13
}

\maketitle

\begin{abstract}
Statistical language models frequently suffer from a lack of training data.
This problem can be alleviated by  clustering,
because it reduces the number of free parameters that need to be trained.
However, clustered models have the following drawback:
if there is ``enough'' data to train an unclustered model, then the clustered
variant may perform worse. On currently used
language modeling corpora, e.g. the Wall Street Journal corpus, how do the
performances of a clustered and an
unclustered model compare? While trying to address this question, we develop
the following two ideas. First, to get
a clustering algorithm with potentially high performance, an existing algorithm
is extended to deal with higher order N-grams.
Second, to make it possible to cluster large amounts of training data more
efficiently,  a heuristic to speed up the
algorithm is presented. The resulting clustering algorithm can be used to
cluster trigrams on the Wall Street Journal corpus and the
language models it produces can compete with existing back-off models.
Especially when there is only little training data available, the clustered
models clearly outperform the back-off models.

\pagebreak
\end{abstract}

\section{Introduction}

It is well known that statistical language models often suffer from a lack of
training data. This is true for standard tasks and
even more so when one tries to build a language model for a new domain, because
a large corpus of texts from that domain is usually
not available. One frequently used approach to alleviate this problem is to
construct a clustered language model. Because it
has fewer parameters, it needs less training data. The main advantage of a
clustered model are its robustness, even in the
face of little or sparse training data,  and its compactness. Particularly when
a language model is used during the acoustic
search in a speech recogniser, having a more compact, e.g. less complex model,
can be of considerable importance. The main drawback
of clustered models is that they may perform worse than an unclustered model,
if there is ``enough'' data to train the
latter. Do corpora currently used for language modeling, e.g. the Wall Street
Journal corpus,  contain enough data in that sense?
Or, in other words, how does the performance of a clustered model compare with
that of an unclustered model? In this paper, we
will attempt to partly answer this question and, along the way, an extended,
more efficient clustering
algorithm will be developed.

In the next section (section \ref{background}), a brief review of existing
clustering algorithms will be given. For the work presented here, we use the
clustering
algorithm proposed in \cite{Kne93}, because, in the spirit of decision directed
learning, it uses an optimisation function
that is very closely related to the final performance measure we wish to
maximise. Since the algorithm forms the basis of our
work, its optimisation criterion is derived in detail.

In order to achieve a clustered model with potentially high
performance, the algorithm is then extended (section \ref{extension})  so that
it can cluster higher order N-grams. We
present three possible approaches for this extension and then develop the one
chosen
for this work in more detail.

When such a clustering algorithm is applied to a large training corpus, e.g.
the Wall Street Journal corpus, with tens of
millions of words, the computational effort required can easily become
prohibitive. Therefore, a simple
heuristic to speed up the algorithm is developed in section \ref{heuristic}.
Its main idea is as follows. Rather than trying to move each word $w$
to all possible
clusters, as the algorithm requires initially, one only tries moving $w$ to a
fixed number of clusters that have been
selected from all possible clusters by a simple heuristic. This reduces the
order of the complexity of the
algorithm. Of course, it may lead to a decrease in performance. However, in
practice, the decrease in performance is
minor (less than 5\%), whereas the obtained speedup is large (up to a factor
of 32).

Because of the increase in the speed of the algorithm, it can be applied more
easily to the Wall Street Journal
corpus and the obtained results will be presented in section \ref{results}.

\section{Background and Related Work}
\label{background}

In speech recognition, one is given a sequence of acoustic observations $A$ and
one tries to find the word
sequence $W^{*}$ that is most likely to correspond to $A$. In order to minimise
the average probability of
error, one should, according to Bayes' decision rule (\cite[p.17]{Dud73}),
choose
\begin{equation}
W^{*}=argmax_{W} p(W|A). \label{eq:basic}
\end{equation}
Based on Bayes' formula (see for example \cite[p.150]{Fre88}), one can rewrite
the probability from the right hand side of equation \ref{eq:basic} according
to the
following equation:
\begin{equation}
p(W|A)=\frac{p(W)*p(A|W)}{p(A)}.
\end{equation}
$p(W)$ is the probability that the word sequence $W$ is spoken, $p(A|W)$ is
the conditional probability that the acoustic signal $A$ is observed when $W$
is spoken
and $p(A)$ is the probability of observing the acoustic signal $A$. Based on
this formula, one can rewrite the maximization of equation \ref{eq:basic} as
\begin{equation}
W^{*}=argmax_{W} \frac{p(W)*p(A|W)}{p(A)} \label{eq:orig_max}.
\end{equation}
Since $p(A)$ is the same for all  $W$, the factor $p(A)$ does not influence the
choice of $W$ and maximising equation \ref{eq:orig_max} is equivalent to
maximising
\begin{equation}
W^{*}=argmax_{W} p(W)*p(A|W) \label{eq:max}.
\end{equation}
The component of the speech recogniser that calculates
$p(A|W)$ is called the acoustic model, the component calculating $p(W)$
the language model. With $W=w_{1},...,w_{n}$, one can further decompose $p(W)$
using the definition of conditional
probabilities as
\begin{equation}
p(W)=\ \prod_{i=1}^{i=n} p(w_{i}|w_{1},...,w_{i-1}) \label{eq:mult}.
\end{equation}
In practice, because of the large number of parameters  in equation
\ref{eq:mult}, the probability of $w_{i}$
usually only depends on the immediately preceding $M$ words:
\begin{equation}
p(w_{i}|w_{1},...,w_{i-1}) \approx p(w_{i}|w_{i-M},...,w_{i-1})
\label{eq:mgram}.
\end{equation}
These models are called $(M+1)$-gram models and in practice, mostly bigram
($M=1$) and trigram ($M=2$) models
are used. Even in these cases, the number of parameters that need to be
estimated from training
data can be quite large. For a speech recogniser with a vocabulary of $20,000$
words, the bigram
needs to estimate roughly $20,000^{2}=4*10^{8}$ parameters and a trigram
$20,000^{3}=8*10^{12}$.

One way to alleviate this problem is to use class based models. Let $G: w
\rightarrow G(w)=g_{w}$ be a function that maps each
word $w$ to its class $G(w)=g_{w}$ and let $|G|$ denote the number of classes.
We can then model the probability
of $w_{i}$ as
\begin{eqnarray}
p(w_{i}|w_{1},...,w_{i-1})& \approx & p_{G}(w_{i}|w_{i-M},...,w_{i-1})
\label{eq:class}\\
 & = & p_{G}(G(w_{i})|G(w_{i-M}),...,G(w_{i-1}))*p_{G}(w_{i}|G(w_{i}))
\label{eq:class2}.
\end{eqnarray}
Thus, if $|G|$=1000 classes are being used, the class-based bigram model
\footnote{This model is also sometimes referred to as bi-pos model, where
pos stands for Parts Of Speech.}
has
$1,000^{2}+20,000=1.02*10^{6}$ parameters and the class-based trigram model
$1,000^{3}+20,000=1.00002*10^{9}$.
This constitutes a significant reduction in both cases.

Many researchers have developed algorithms for determining the clustering
function $G$ automatically
(see for example \cite{Car94}, \cite{Jar93}, \cite{Jel90}, \cite{Kne93} and
\cite{Per93}). Starting from
an initial clustering function, the basic principle often is to move words from
their current cluster to
another cluster, but only if this improves the value of an optimisation
criterion. The algorithms often
differ in the optimisation criterion and in general, there
are many possible choices for it. However, in the spirit of decision-directed
learning, it makes sense to
use as optimisation criterion a function that is very closely related or
identical to the final
performance measure we wish to maximise. This way, one can be very confident
that an improvement in the
optimisation criterion will actually translate to an improvement of
performance. We therefore chose the
algorithm proposed in \cite{Kne93} as the basis for our work. In the following,
the
optimisation criterion for a bigram based model (e.g. $M=1$) will be
derived, roughly as presented in \cite{Kne93}.

In order to automatically find classification functions  $G$, the
classification problem is first converted into an optimisation problem.
Suppose the function $F(G)$ indicates how good the classification $G$  is. One
can then
reformulate the classification problem as finding the classification $G^{*}$
that maximises F:
\begin{equation} G^{*} = argmax_{G \in {\cal G}} F(G),
\end{equation}
where ${\cal G}$ contains the set of possible classifications which are at our
disposal.

What is a suitable function $F$, also called optimisation criterion? Given a
classification function
$G$, the probabilities $p_{G}(w|v)$ of equation \ref{eq:class2} can be
estimated using
the maximum likelihood (ML) estimator, e.g. relative frequencies:
\begin{eqnarray}
p_{G}(w|v) & = & p(G(w)|G(v)) * p(w|G(w))\\
 & = & \frac{N(G(v), G(w))}{N(G(v))} * \frac{N(G(w), y)}{N(G(w))},
\end{eqnarray}
where $N(x)$ denotes the number
of times $x$ occurs in the training data.
Given these probability estimates $p_{G}(w|v)$, the likelihood $F_{ML}$
of the training data, e.g. the probability of the training data being generated
by our probability
estimates $p_{G}(w|v)$, measures how well the training data is represented by
the
estimates and can be used as optimisation criterion (\cite{Jel90}).
The likelihood of the
training data $F_{ML}$ is simply
\begin{eqnarray}
F_{ML} & = & \prod_{i=1}^{N} p_{G}(w_{i}|w_{i-1})\\
 & = &  \prod_{i=1}^{N} \frac{N(g_{w_{i-1}}, g_{w_{i}})}{N(g_{w_{i-1}})} *
\frac{N(g_{w_{i}}, w_{i})}{N(g_{w_{i}})}.
\end{eqnarray}
Assuming that the classification is unique, e.g. that $G$ is a function,
$N(g_{w_{i}}, w_{i})=N(w_{i})$ always holds
(because $w_{i}$ always occurs with the same class $g_{w_{i}}$). Since one
tries to optimise $F_{ML}$ with respect to $G$, any term that does not depend
on $G$
can be removed, because it will not influence the optimisation. It is
thus equivalent to optimise
\begin{eqnarray}
F^{'}_{ML} & = & \prod_{i=1}^{N} \frac{N(g_{w_{i-1}},
g_{w_{i}})}{N(g_{w_{i-1}})} * \frac{1}{N(g_{w_{i}})}\\
 & =: & \prod_{i=1}^{N} f(w_{i-1}, w_{i}).
\end{eqnarray}
If, for two pairs $(w_{i-1}, w_{i})$ and $(w_{j-1}, w_{j})$,
$G(w_{i-1})=G(w_{j-1})$ and
$G(w_{i})=G(w_{j})$ holds, then $f(w_{i-1}, w_{i})=f(w_{j-1}, w_{j})$ is also
true.
Identical terms can thus be regrouped to obtain
\begin{equation}
 F^{'}_{ML}= \prod_{g_{1}, g_{2}} [ \frac{N(g_{1}, g_{2})}{N(g_{1})} *
\frac{1}{N(g_{2})} ] ^{N(g_{1}, g_{2})},
\end{equation}
where the product is over all possible pairs $(g_{1}, g_{2})$.
Because $N(g_{1})$ does not depend on $g_{2}$ and $N(g_{2})$ does not depend on
$g_{1}$, this can again be simplified to
\begin{equation}
F^{'}_{ML}= \prod_{g_{1}, g_{2}} N(g_{1}, g_{2})^{N(g_{1}, g_{2})}
\prod_{g_{1}} \frac{1}{N(g_{1})}^{N(g_{1})}
\prod_{g_{2}} \frac{1}{N(g_{2})}^{N(g_{2})}.
\end{equation}
After taking the logarithm, one obtains the equivalent optimisation criterion
$F^{''}_{ML}$
\begin{eqnarray}
 F^{''}_{ML} & = &  \sum_{g_{1}, g_{2}}  N(g_{1}, g_{2}) * log( N(g_{1}, g_{2}
)) -
\sum_{g_{1}} N(g_{1}) * log ( N(g_{1}))\\
 &  - &  \sum_{g_{2}} N(g_{2}) * log ( N(g_{2})) \nonumber .
\end{eqnarray}

$F^{''}_{ML}$ is the maximum likelihood optimisation criterion that can be used
 to find a good classification
$G$. However, the problem with this maximum likelihood criterion is that one
first estimates the probabilities
$p_{G}(w|v)$ on the training data $T$ and then, given $p_{G}(w|v)$, one
evaluates
the classification $G$ on $T$. In other words, both the classification $G$ and
the estimator
$p_{G}(w|v)$ are trained on the same data. Thus, there will not be any unseen
event, a fact
that overestimates the power for generalisation of the class based model. In
order to avoid this,
a cross-validation technique will be incorporated directly into the
optimisation criterion in section \ref{lo}.

\subsection{Leaving-One-Out Criterion}
\label{lo}

The basic principle of cross-validation is to split the training data $T$ into
a ``retained'' part
$T_{R}$ and a ``held-out'' part $T_{H}$. One can then use $T_{R}$ to estimate
the probabilities
$p_{G}(w|v)$ for a given classification $G$,  and $T_{H}$  to evaluate how well
the
classification $G$ performs. The so-called
leaving-one-out technique is a special case of cross-validation
(\cite[pp.75]{Dud73}).
It divides the data into $N-1$ samples as ``retained'' part and only one sample
as ``held-out'' part. This is repeated $N-1$ times, so that each sample is
once in the ``held-out'' part. The advantage of this approach is that all
samples
are used in the ``retained'' and in the ``held-out'' part, thus making very
efficient
use of the existing data. In other words, the  ``held-out'' part $T_{H}$ to
evaluate
a classification $G$ is the entire set of data points; but when we calculate
the
probability of the $i^{th}$ data point, one assumes that the probability
distributions
$p_{G}(w|v)$ were estimated on all the data expect point $i$.

Let $T_{i}$ denote the data without the pair $(w_{i-1}, w_{i})$ and
$p_{G,T_{i}}(w|v)$ the probability estimates based on a given classification
$G$ and
training corpus $T_{i}$. Given a particular $T_{i}$, the probability of the
``held-out''
part $(w_{i-1}, w_{i})$ is $p_{G,T_{i}}(w_{i}|w_{i-1})$. The probability of the
complete corpus,
where each pair is in turn considered the ``held-out'' part is the
leaving-one-out likelihood $L_{LO}$
\begin{equation}
 L_{LO}=\prod_{i=1}^{N} p_{G,T_{i}}(w_{i}|w_{i-1}).
\label{eq:two}
\end{equation}
In the following, an optimisation function $F_{LO}$ will be derived by
specifying how
$p_{G,T_{i}}(w_{i}|w_{i-1})$ is estimated from frequency counts.
 First $p_{G,T_{i}}(w_{i}|w_{i-1})$ is rewritten  as usual (see equation
\ref{eq:class2}):
\begin{eqnarray}
 p_{G,T_{i}}(w|v) & = & p_{G,T_{i}}(G(w)|G(v))*p_{G, T_{i}}(w|G(w))\\
 & = & \frac{p_{G,T_{i}}(g_{1}, g_{2})}{p_{G,T_{i}}(g_{1})} *
\frac{p_{G,T_{i}}(g_{2}, w)}{p_{G,T_{i}}(g_{2})},
\label{eq:estim}
\end{eqnarray}
where $g_{1}=G(v)$ and $g_{2}=G(w)$.
Now we will specify how each term in equation \ref{eq:estim} is estimated.

As shown before, $p_{G,T_{i}}(g_{2}, w)=p_{G,T_{i}}(w)$ (if the classification
$G$
is a function) and since $p_{T_{i}}(w)$ is actually independent of $G$, one can
drop it
out of the maximization and thus need not specify an estimate for it.

As will be shown later, one can guarantee that every class
$g_{1}$ and $g_{2}$ has been seen at least once in the
``retained'' part and one can thus use relative counts as estimates for class
uni-grams:
\begin{eqnarray}
p_{G,T_{i}}(g_{1}) & = & \frac{N_{T_{i}}(g_{1})}{N_{T_{i}}} \label{eq:three}\\
p_{G,T_{i}}(g_{2}) & = & \frac{N_{T_{i}}(g_{2})}{N_{T_{i}}} \label{eq:four} .
\end{eqnarray}

However, in the case of the class bi-gram, one might have to predict unseen
events
\footnote{If $(w_{i-1}, w_{i})$ occurs only once in the complete corpus, then
$p_{G,T_{i}}(w_{i}|w_{i-1})$ will have to be calculated based on the corpus
$T_{i}$, which does not
contain any occurrences of $(w_{i-1}, w_{i})$.}.
We therefore use the absolute discounting
method (\cite{Ney93}), where the counts are reduced by a constant value $b < 1$
and where the
gained probability mass is redistributed over unseen events. Let
$n_{0,T_{i}}$ be the number of unseen pairs $(g_{1}, g_{2})$ and $n_{+,T_{i}}$
the number of
seen pairs $(g_{1}, g_{2})$. This leads to the following smoothed estimate
\begin{eqnarray}
\lefteqn{ p_{G,T_{i}}(g_{1}, g_{2})} \nonumber \\
 & = & \left\{ \begin{array}{ll}
\frac{N_{T_{i}}(g_{1}, g_{2}) - b}{N_{T_{i}}} & \mbox{if $N_{T_{i}}(g_{1},
g_{2})>0$}\\
\frac{n_{+, T_{i}}*b}{n_{0,T_{i}}*N_{T_{i}}} & \mbox{if $N_{T_{i}}(g_{1},
g_{2})=0$}\\
\end{array}
\right.
\end{eqnarray}
Ideally, one would make $b$ depend on the classification, e.g. use
$b=\frac{n_{1}}{n_{1}+2*n_{2}}$, where
$n_{1}$ and $n_{2}$ depend on $G$. However, due to computational reasons, we
use, as suggested in
\cite{Kne93}, the empirically determined constant value $b=0.75$ during
clustering.
The probability distribution $p_{G,T_{i}}(g_{1}, g_{2})$ will always be
evaluated on the
``held-out'' part $(w_{i-1}, w_{i})$ and with $g_{1,i}=g_{w_{i-1}}$ and
$g_{2,i}=g_{w_{i}}$, one obtains
\begin{eqnarray}
\lefteqn{ p_{G,T_{i}}(g_{1,i}, g_{2,i})} \nonumber \\
 & = & \left\{ \begin{array}{ll}
\frac{N_{T_{i}}(g_{1,i}, g_{2,i}) - b}{N_{T_{i}}} & \mbox{if
$N_{T_{i}}(g_{1,i}, g_{2,i})>0$}\\
\frac{n_{+, T_{i}}*b}{n_{0,T_{i}}*N_{T_{i}}} & \mbox{if $N_{T_{i}}(g_{1,i},
g_{2,i})=0$}
\label{eq:five}
\end{array}
\right.
\end{eqnarray}

In order to facilitate future regrouping of terms, one can now express the
counts
$N_{T_{i}}, N_{T_{i}}(g_{1})$ etc.
in terms of the counts of
the complete corpus $T$ as follows:
\begin{eqnarray}
N_{T_{i}} & = & N_{T} - 1 \label{eq:six} \\
N_{T_{i}}(g_{1}) & = & N_{T}(g_{1}) - 1 \\
N_{T_{i}}(g_{2}) & = & N_{T}(g_{2}) - 1 \\
N_{T_{i}}(g_{1,i}, g_{2,i}) & = &  N_{T}(g_{1,i}, g_{2,i})-1\\N_{T_{i}} & = &
N_{T} - 1 \\
n_{+, T_{i}} & = & \left\{
\begin{array}{ll}
n_{+, T} & \mbox{if $N_{T}(g_{1,i}, g_{2,i})>1$}\\
n_{+, T} - 1  & \mbox{if $N_{T}(g_{1,i}, g_{2,i})=1$}\\
\end{array} \right. \\
n_{0,T_{i}} & = &  \left\{
\begin{array}{ll}
n_{0,T} & \mbox{if $N_{T}(g_{1,i}, g_{2,i})>1$}\\
n_{0, T} - 1  & \mbox{if $N_{T}(g_{1,i}, g_{2,i})=1$}
\label{eq:sixb}
\end{array} \right.
\end{eqnarray}
All the expressions can now be substituted back into equation \ref{eq:two}.
After dropping $p_{G,T_{i}}(w)$ because it is independent of $G$, one arrives
at
\begin{eqnarray}
F'_{LO} & = & \prod_{i=1}^{N} \frac{p_{G,T_{i}}(g_{1,i},
g_{2,i})}{p_{G,T_{i}}(g_{1,i})} * \frac{1}{p_{G,T_{i}}(g_{2,i})}\\
& = & \prod_{g_{1}, g_{2}} ( p_{G,T_{i}}(g_{1}, g_{2}))^{N(g_{1}, g_{2})} *
\prod_{g_{1}} (\frac{1}{p_{G,T_{i}}(g_{1})})^{N(g_{1})}\\
&  & * \prod_{g_{2}} (\frac{1}{p_{G,T_{i}}(g_{2})})^{N(g_{2})} \nonumber .
\end{eqnarray}
One can now substitute equations \ref{eq:three}, \ref{eq:four} and
\ref{eq:five}, using
the counts of the whole corpus of equations \ref{eq:six} to \ref{eq:sixb} .
After having dropped
terms independent of $G$, one obtains
\begin{eqnarray}
F''_{LO} & = & \prod_{g_{1}, g_{2} : N(g_{1},g_{2})  > 1 } (N_{T}(g_{1}, g_{2})
-1 -b )^{N_{T}(g_{1}, g_{2})} *
\left( \frac{(n_{+,T}-1)*b}{(n_{0,T}+1)} \right)^{n_{1,T}}\\
 &  * & \prod_{g_{1}} \left( \frac{1}{(N_{T}(g_{1}-1))} \right)^{N_{T}(g_{1})}
* \prod_{g_{2}} \left( \frac{1}{(N_{T}(g_{2}-1))} \right)^{N_{T}(g_{2})}
\nonumber ,
\end{eqnarray}
where $n_{1,T}$ is the number of pairs $(g_{1}, g_{2})$ seen exactly once in
$T$
(e.g. the number of pairs that will be unseen when used as ``held-out'' part).
Taking the logarithm, we obtain the final optimisation criterion $F'''_{LO}$
\begin{eqnarray}
F'''_{LO} & = & \sum_{g_{1}, g_{2}:N_{T}(g_{1}, g_{2})>1} N_{T}(g_{1}, g_{2}) *
log ( N_{T}(g_{1}, g_{2}) - 1 - b )\\
 & + & n_{1,T} * log ( \frac{b*(n_{+, T}-1)}{(n_{0,T}+1)} ) \nonumber \\
 & - & \sum_{g_{1}} N_{T}(g_{1}) * log (  N_{T}(g_{1}) - 1 ) - \sum_{g_{2}}
N_{T}(g_{2}) * log (  N_{T}(g_{2}) - 1 )
\nonumber .
\end{eqnarray}

\subsection{Clustering Algorithm}
\label{algo}

Given the  maximization criterion $F'''_{LO}$, we use the algorithm in Figure
\ref{fig:algo} to find a good clustering
function $G$. The algorithm tries to make local changes by moving words between
classes, but only if
it improves the value of the optimisation function. The algorithm will converge
because the
optimisation criterion is made up of logarithms of probabilities and thus has
an upper limit and
because the value of the optimisation criterion increases in each iteration.
However, the solution
found by this greedy algorithm is only locally optimal and it depends on the
starting conditions.
Furthermore, since the clustering of one word affects the future clustering of
other words, the
order in which words are moved is important. As suggested in \cite{Kne93}, the
words
are sorted by
the number of times they occur such that the most frequent words, about which
one knows
the most,  are clustered first. Moreover, infrequent words (e.g. words
with occurrence counts smaller than $5$) are not considered for clustering,
because the information they
provide is not very reliable. Thus, if one starts out with an initial
clustering in which no cluster
occurs only once, and if one never moves words that occur only once,
then one will never have a cluster which occurs only once.
Thus, the assumption we made earlier,
when it was decided to estimate cluster uni-grams by frequency counts, can be
guaranteed.
\begin{figure}
\begin{algorithm}{Clustering}{}
\begin{listbl}
	\item  start with initial clustering function $G$
	\item iterate until some convergence criterion is met
	\item \{
	\begin{listbl}
		\item for all $w \in V$
		\item \{
		\begin{listbl}
			\item for all $g'_{w} \in G$
			\item \{
			\begin{listbl}
				\item calculate the difference in $F(G)$ when $w$ is moved from $g_{w}$ to
$g'_{w}$
			\end{listbl}
			\item \}
			\item move $w$ to $g'_{w}$ that results in biggest improvement in $F(G)$
		\end{listbl}
		\item \}
	\end{listbl}
	\item \}
\end{listbl}
\end{algorithm}
\caption{The clustering algorithm}
\label{fig:algo}
\end{figure}

We will now determine the complexity of the algorithm. Let $C$ be
the maximal number of clusters for $G$, let $E$ be the number
of elements one tries to cluster (e.g. $E=|V|$),
and let $I$ be the number of iterations.
When one moves $w$ from $g_{w}$ to $g'_{w}$ in the inner loop, one needs
to change the counts $N(g_{w}, g_{2})$ and $N(g'_{w}, g_{2})$ for all $g_{2}$.
The amount by which the counts need to be changed is equal to the
number of times $w$ occurred with cluster $g_{2}$. Since this amount is
independent of $g'_{w}$, one only needs to calculate it once for each $w$.
The amount can then be looked up in constant time within the loop, thus
making the inner loop of order $C$. The inner loop is executed once for every
cluster $w$ can be moved to, thus giving a complexity of the order of $C^{2}$.
For
each $w$, one needed to calculate the number of times $w$ occurred with all
clusters $g_{2}$. For that one has to sum up all the bigram counts
$N(w,v):G(v)=g_{2}$, which is on the order of $E$, thus giving a complexity of
the order
of  $E+C^{2}$. The two outer loops are executed $I$ and $E$ times, thus giving
a total complexity of the order of $I*E*(E+C^{2})$.

\section{Extending the Clustering Algorithm to $N$-grams}
\label{extension}

It is well known that a trigram model outperforms a bigram model if there is
sufficient training data. If
we want our clustering algorithm to compete with unclustered models on a corpus
like the Wall Street Journal,
where the trigram indeed outperforms the bi-gram, it therefore seems logical
that the
clustering algorithm should be extended to deal with trigrams (and higher order
$N$-grams) as well.
The original clustered bigram model, as derived from equation \ref{eq:class2},
is
\begin{eqnarray}
p(w_{i}|w_{i-1})& = & p_{G}(G(w_{i})|G(w_{i-1}))*p_{G}(w_{i}|G(w_{i}))
\label{eq:biclass}.
\end{eqnarray}
There are at least three ways of extending the clustering to $(M+1)$-grams,
depending on how one models
the probability $p(w_{i}|w_{i-M},...,w_{i-1})$:
\begin{eqnarray}
a) & & p_{G}(G(w_{i})|G(w_{i-M}),...,G(w_{i-1}))*p_{G}(w_{i}|G(w_{i})) \\
b) & &
p_{G}(G_{M+1}(w_{i})|G_{M}(w_{i-M}),...,G_{1}(w_{i-1}))*p_{G}(w_{i}|G_{M+1}(w_{i}))\\
c) & & p_{G}(G_{2}(w_{i})|G_{1}(w_{i-M},...,w_{i-1}))*p_{G}(w_{i}|G_{2}(w_{i}))
\label{eq:extending}
\end{eqnarray}

The tradeoff between these models is one of accuracy versus complexity.
Approach $a)$, which only
uses one clustering function $G$, could produce
\[
 |G|^{|V|}
\]
different clusterings (for each word in $V$,
it can choose one of the $|G|$ clusters). Approach $b)$, which uses $M+1$
different clustering functions,
can represent
\[
\sum_{i=0}^{i=M+1} |G_{i}|^{V}
\]
different clusterings, including all the clusterings of
approach $a)$. Approach $c)$, which uses one clustering function for the tuples
$w_{i-M},...,w_{i-1}$ and
one for $w_{i}$, can produce
\[
|G_{1}|^{(|V|^{M})}+|G_{2}|^{|V|}
\]
possible clusterings, including all the
ones represented by approach $a)$ and $b)$. Approach $c)$ therefore has the
highest potential for accuracy,
as long as there is sufficient training data. Since the Wall Street Journal
corpus is very large, we
decided to use approach $c)$. Please note that for $M=1$, approach $a)$ gives
the traditional
clustered bigram approach, but approaches $b)$ and $c)$ ($c)$ collapses to $b)$
for $M=1$) are more
general than the traditional model. Moreover, approach $c)$ is referred to in a
recent publication
(\cite{Ney94}) as a two-sided (non symmetric) approach.

Similar to the derivation presented in section \ref{background}, one can now
derive the  optimisation
criterion for approach $c)$. However,  since it is very similar to the
derivation shown in section \ref{background}, only the final formulae will be
given here. The complete derivation is given in appendix A.
Let $g_{1}$ and $g_{2}$ denote clusters of $G_{1}$ and $G_{2}$ respectively.
The optimisation criterion for the extended algorithm is
\begin{eqnarray}
F_{LO} & = & \sum_{g_{1}, g_{2}:N_{T}(g_{1}, g_{2})>1} N_{T}(g_{1}, g_{2}) *
log ( N_{T}(g_{1}, g_{2}) - 1 - b )\\
 & + & n_{1,T} * log ( \frac{b*(n_{+, T}-1)}{(n_{0,T}+1)} ) \nonumber \\
 & - & \sum_{g_{1}} N_{T}(g_{1}) * log (  N_{T}(g_{1}) - 1 ) - \sum_{g_{2}}
N_{T}(g_{2}) * log (  N_{T}(g_{2}) - 1 )
\nonumber .
\end{eqnarray}
The corresponding clustering algorithm, which is shown in figure
\ref{fig:algob}, is a straight forward extension of the
one given in section \ref{background}.
\begin{figure}
\begin{algorithm}{Clustering}{}
\begin{listbl}
	\item  start with initial clustering function $G$
	\item iterate until some convergence criterion is met
	\item \{
	\begin{listbl}
		\item for all $w \in V$ and $t \in V^{M}$
		\item \{
		\begin{listbl}
			\item for all $g'_{w} \in G_{2}$ and $g'_{t} \in G_{1}$
			\item \{
			\begin{listbl}
				\item calculate the difference in $F(G)$ when $w$/$t$ is moved from
$g_{w}$/$g_{t}$ to $g'_{w}$/$g'_{t}$
			\end{listbl}
			\item \}
			\item move the $w$/$t$ to the $g'_{w}$/$g'_{t}$ that results in the biggest
improvement in $F(G)$
		\end{listbl}
		\item \}
	\end{listbl}
	\item \}
\end{listbl}
\end{algorithm}
\caption{The extended clustering algorithm}
\label{fig:algob}
\end{figure}
It's complexity can be derived as follows. Let $C_{G_{1}}$ and $C_{G_{2}}$ be
the maximal number of clusters for $G_{1}$ and $G_{2}$, let $E_{1}$ and $E_{2}$
 be the number
of elements $G_{1}$ and $G_{2}$ try to cluster, let $C=max(C_{G_{1}},
C_{G_{2}})$,
$E=max(E_{1}, E_{2})$ and let $I$ be the number of iterations.
When one moves $w$ from $g_{w}$ to $g'_{w}$ in the inner loop
(the situation is symmetrical for $t$), one needs
to change the counts $N(g_{w}, g_{2})$ and $N(g'_{w}, g_{2})$ for all $g_{2}
\in G_{2}$.
The amount by which the counts need to be changed is equal to the
number of times $w$ occurs with cluster $g_{2}$. Since this amount is
independent of $g'_{w}$, one only needs to calculate it once for each $w$.
The amount can then be looked up in constant time within the loop, thus
making the inner loop of order $C$. The inner loop is executed once for every
cluster $w$ can be moved to, thus giving a complexity of the order of $C^{2}$.
For
each $w$, one needed to calculate the number of times $w$ occurred with all
clusters $g_{2}$. For that, one has to sum up all the bigram counts
$N(w,t):G_{2}(t)=g_{2}$, which is on the order of $E$, thus giving a complexity
of the order
of  $E+C^{2}$. The two outer loops are executed $I$ and $E$ times thus giving
a total complexity of the order of $I*E*(E+C^{2})$. This is almost identical
to the complexity of the bigram clustering algorithm given in section
\ref{background}, except that $E$ is now the number of $(M+1)$-grams one wishes
to cluster, rather than the number of unigrams (e.g. words of the vocabulary).

\section{Speeding up the Algorithm}
\label{heuristic}

If one wants to use the clustering algorithm on large corpora, the complexity
of the algorithm
becomes a crucial issue. As shown in the last two sections, the complexity of
the algorithm
is $O(I*E*(E+C^{2}))$, where $C$ is the maximally allowed number of clusters,
$I$ is the number of
iterations and $E$ is the number of elements to cluster ($|V|$ in case of
bigrams, $|V|^{M+1}$ in case of the extended algorithm). $C$ crucially
determines the quality of the resulting language model and one
would therefore like to chose it as big as possible. Unfortunately, because the
algorithm
is quadratic in $C$, this may be very costly. We therefore developed the
following heuristic to speed up the algorithm.

The factor $C^{2}$ comes from the fact that one tries to move a word $w$ to
each of the $C$ possible
clusters ($O(C)$), and for each of these one has to calculate the difference in
the
optimisation function ($O(C)$ again). If, based on some heuristic, one could
select a fixed
number $t$ of target clusters, then one could only try moving $w$ to these $t$
clusters, rather than
to all possible clusters $C$. This may of course lead to the situation where
one does not move a
word to the {\em best} possible cluster (because it was not selected by the
heuristic), and thus
potentially to a decrease in performance. But this decrease in performance
depends of course
on the heuristic function used and we will come back to this issue when we look
at the practical
results.

The heuristic used in this work is as follows. For each cluster $g_{1}$, one
keeps track of the $h$ clusters that
most frequently co-occur with $g_{1}$ in the tables $N(g_{1}, g_{2})$.
For example, if $g_{1}$ is a cluster of $G_{1}$ (the situation is symmetric
for $G_{2}$), then the $h$ biggest entries in $N(g_{1}, g)$ are the
$h$ clusters being stored.
When one tries to move a word $w$,
one also constructs a list of the $h$ most frequent clusters that follow $w$
(one can  get this
for free as part of the factor $E$ in $(E+C^{2})$). One then simply calculates
the
number of clusters that are in both lists and takes this as the heuristic score
$H(g_{1})$. The
bigger $H(g_{1})$, the more similar are the distributions of $w$ and $g_{1}$,
and the more likely
it is that $g_{1}$ is a good target cluster to which $w$ should be moved.
Because the
length of the lists is a constant $h$  calculating the heuristic score
is also independent of $C$. One can thus calculate the heuristic score of all
$C$ clusters in $O(C)$.
However, once one has decided to move $w$ to a given cluster, one would have to
update the
lists containing the $h$ most frequent clusters following each cluster $g_{1}$
(the lists might have
changed due to the last moving of a word). Since the update is $O(C)$ for a
given $g_{1}$, the update
would again be $O(C^{2})$ for all clusters. In order to avoid this, one can
make another
approximation at this point. One can only update the list for the original  and
the new cluster of $w$. The
full update of all the lists  is only performed after a certain number $u$ of
words have been moved.

To sum up, we can say that one can select $t$ target clusters using the
heuristic in $O(C)$. Following
that, one tries  moving $w$ to each of these $t$ clusters, which is again
$O(C)$. Moreover, several times
per iteration (depending on $u$), one updates the list of most frequent
clusters which is $O(C^{2})$.
Thus, the complexity of the heuristic version of the algorithm is
$O(I*(E*(E+C)+C^{2}))$. The complexity still contains the factor $C^{2}$, but
this time not
within the inner parenthesis. The factor $C^{2}$ will thus be smaller than
$E*(E+C)$, and is only given for completeness.

We will now present a practical evaluation of the heuristic algorithm. The
heuristic itself is
parameterised by $h$, the number of most frequent clusters one uses to
calculate the heuristic score, $t$,
the number of best ranked target clusters one tries to move word $w$ to and
$u$, the number indicating after how many words a full update of the list of
most frequent clusters
is performed. In order to evaluate the heuristic for a given set
of parameters, one can simply compare the final value of the approximation
function
and the resulting perplexity of the
heuristic algorithm with that of the full algorithm.

Table \ref{tab:data_baseline} contains a comparison of the results using
approximately one million words of training data (from the Wall Street
Journal corpus) and values $t=10$, $h=10$ and
$u=1000$. The CPU Time given was measured on a DEC alpha workstation (DEC 3000,
model 600), which
was used in all the experiments reported in this paper.
One can see that the
execution time of the standard algorithm seems indeed quadratic in the number
of
clusters, whereas that of the heuristic version seems to be linear. Moreover,
the perplexity of the heuristic version is always within 4\% of that of the
standard algorithm, a number which does not seem to increase systematically
with
the number of clusters. Furthermore, the speed up of the algorithm seems to
be closely related to the number of clusters divided by $t$. For example,
in the case of $320$ clusters, this ration is $320/10=32$ and the heuristic
version is indeed almost $32$ times as fast (the speed up is almost
$1-\frac{1}{32}=0.97$).
Judging from the time behaviour of the standard algorithm, one would expect
it to take around 32 hours to run with $1000$ clusters, whereas the
heuristic algorithm, as will be shown later, only takes about half an
hour  (for $t=10$).
\begin{table}
\centering
\begin{tabular}{|c|cc|cc|} \hline
Clusters  & \multicolumn{2}{c|}{Standard Algorithm} &
\multicolumn{2}{c|}{Heuristic Version}\\
 & PP & CPU Time & PP ($\Delta$ \%) & CPU Time ($\Delta$\%)\\ \hline
10	&746	&1:02	&746 (0.0)	&1:05 (4.8)\\
20	&630	&2:28	&653 (3.6)	&1:35 (-36)\\
40	&548	&7:46	&558 (1.8)	&2:36 (-67)\\
80	&477	&26:41	&490 (2.7)	&4:30 (-83)\\
160	&421	&1:33:29 &437 (3.8)	&7:59 (-91)\\
320	&394	&5:20:18 &402 (2.0)	&14:17 (-96)\\ \hline
\end{tabular}
\caption{Comparison of the algorithm with its heuristic version}
\label{tab:data_baseline}
\end{table}

Tables \ref{tab:data_t} to \ref{tab:data_h} contain a more detailed analysis
of the influence of the parameters $t$, $u$, and $h$ on the heuristic version
of the algorithm, this time with a maximal number of allowed clusters of
$1000$.
The first point to note is
that in all tables, a change in the value of the optimisation function is very
closely related to a change in perplexity. This is a very reassuring finding,
because it indicates that the clustering algorithm actually tries to
optimise the correct criterion.

{}From table \ref{tab:data_t}, one can see
that an increase in $t$ leads to an increase  in execution time, but also
 to an increase in performance. This is because as $t$ increases,
the chances of the
heuristic missing the overall best target cluster for a given word $w$
decreases.

In table \ref{tab:data_u}, one can see that the effect of $u$ on the
algorithm is very minor. This could be explained by the fact that even
though the full lists of most frequent clusters are not updated at every move,
the update in clusters $g_{w}$ and $g'_{w}$, which is performed at every
move, contains the most important changes.

Finally, in table \ref{tab:data_h}, one can see that the performance of
the algorithm decreases with an increase in $h$. This in a way counter
intuitive result could be explained by the following hypothesis. If the
suitability of a target cluster is determined by a small number of very
frequently co-occurring clusters, then increasing $h$ could make the heuristic
perform worse, because the effect of the most important clusters is
perturbed by a large number of less important clusters (the heuristic
only counts the number of clusters in both lists and does not weigh them).

Based on the results of these experiments, we chose $h=5$, $t=10$ and
$u=1000$ for future experiments with the heuristic version of the algorithm.
\begin{table}
\centering
\begin{tabular}{|cccc|} \hline
t & opt & PP & time \\ \hline
5	&-1.246e+07	&359		&20:24\\
10	&-1.243e+07	&354		&31:35\\
20	&-1.241e+07	&350		&55:32\\
40	&-1.240e+07	&349		&1:34:16\\
80	&-1.239e+07	&348		&2:51:19\\ \hline
\end{tabular}
\caption{Results for $u=1000$ and $h=10$}
\label{tab:data_t}
\end{table}

\begin{table}
\centering
\begin{tabular}{|cccc|} \hline
u & opt & PP & time \\ \hline
4	&-1.243e+07	&352		&36:27\\
20	&-1.243e+07	&353		&34:00\\
100	&-1.243e+07	&353		&34:00\\
500	&-1.243e+07	&354		&34:00\\
2500	&-1.243e+07	&354		&32:38\\ \hline
\end{tabular}
\caption{Results for $t=10$ and $h=10$}
\label{tab:data_u}
\end{table}

\begin{table}
\centering
\begin{tabular}{|cccc|} \hline
h & opt & PP & time \\ \hline
5	&-1.243e+07	&353		&31:09\\
10	&-1.243e+07	&353		&33:09\\
20	&-1.245e+07	&357		&36:26\\
40	&-1.254e+07	&370		&41:50\\ \hline
\end{tabular}
\caption{Results for $t=10$ and $u=10$}
\label{tab:data_h}
\end{table}

\section{Results}
\label{results}

In the following, we will present  results of clustered language models on  the
Wall
Street Journal (WSJ) corpus. The work reported here was performed on the WSJ0
corpus, using the verbalised pronunciation (VP) and non verbalised
pronunciation
(NVP) versions of the corpus with the 20K open vocabulary. As mentioned
on the CDROM (and as discussed in \cite{Ueb93g}), the results for open
vocabularies are usually not meaningful,
if the unknown words are taken into account when calculating the
perplexity.
One way to solve this problem (\cite{Ros94b}) is to simply skip the
unknown words, when calculating the perplexity. Since all our experiments
were performed with the open vocabulary, this is the approach
taken here, except when indicated otherwise.  In order to investigate
the influence of the amount of training data on the results, we used seven
different sets of training
data, $T1$ to $T7$, with about 2K, 12K, 60K, 350K, 1.7M, 8.5M and 40M words
respectively. All perplexity results
were calculated on approximately 2.3 million words of text that were not part
of the training
material. The clustered models were
produced with the extended heuristic version of the algorithm. To run to
completion, it took less than
12 hours real time for the bigram case, and several days for the trigram case.

As a yardstick for the performance of the clustered models, we implemented the
commonly used compact
back-off model (\cite{Kat89}, \cite{Pau89b}). Because the bigram counts were
not smoothed, the probability
mass, that could be redistributed to unseen events, was only gained through
events that fell below the
cut-off threshold. If a given cut-off threshold did  did not lead to any gained
probability
mass for a particular distribution (because no
event was below the threshold and thus no probability mass could be
redistributed), the cut-off threshold
of this distribution was set to the lowest value, that would lead to some gain
in probability mass.
Table \ref{tab:back_off_vb} and
\ref{tab:back_off_nb} give the perplexity of back-off models with various
cut-off thresholds $C$ for verbalised
and non-verbalised pronunciation respectively.
\begin{table}
\centering
\begin{tabular}{|c|cccc|} \hline
Training Set& $C=250$&	$C=50$&	$C=10$&	$C=2$\\ \hline
T1&	2180&	2180&	2150&	2240\\
T2&	1350&	1320&	1210&	1130\\
T3&	1030&	936&	750&	621\\
T4&	812&	645&	480&	373\\
T5&	620&	462&	330&	254\\
T6&	456&	323&	236&	190\\
T7&	324&	233&	182&	159\\ \hline
\end{tabular}
\caption{Back-off perplexity results on VP data}
\label{tab:back_off_vb}
\end{table}
\begin{table}
\centering
\begin{tabular}{|c|cccc|} \hline
Training Set& $C=250$& $C=50$&	$C=10$&	$C=2$\\ \hline
T1&	3170&	3170&	3150&	3220\\
T2&	1980&	1950&	1790&	1610\\
T3&	1440&	1310&	1060&	878\\
T4&	1170&	936&	696&	537\\
T5&	928&	693&	488&	369\\
T6&	666&	466&	332&	265\\
T7&	464&	327&	250&	216\\ \hline
\end{tabular}
\caption{Back-off perplexity results on NVP data}
\label{tab:back_off_nb}
\end{table}
First, one can see that a bigger value of $C$ leads to a higher perplexity.
This is because as $C$ increases, more
and more bigram counts are discarded and replaced by unigram, rather than
bigram,  probability estimates.
However, a good reason why higher values of $C$ might still be of interest is
that they lead to substantially
smaller models and this can be of crucial importance for the time performance
of a recogniser.
Second, and more importantly for our purposes,
the results seem comparable to other results reported in the literature. In
\cite{Aub94} for example, the perplexity
results for the non-verbalised data with open vocabulary is $205$, quite close
to our $216$ (for $C=2$). However, it
is quite likely that the probabilities of unknown words were taken into account
for the calculation of the
$205$ value and our model also gives a perplexity of $205$ in that case. The
back-off results of tables
\ref{tab:back_off_vb} and \ref{tab:back_off_nb} therefore constitute a
reasonable yardstick to evaluate
the performance of the clustered language models.

Tables \ref{tab:train_pp_vb} and \ref{tab:train_pp_nb} give the results of a
clustered bigram with
$2000$ clusters for both $G_{1}$ and $G_{2}$, for verbalised and non-verbalised
pronunciation respectively.
For better comparison, the matching results of the back-off models are repeated
and the difference is given in percent.
Even though the clustered model performs worse than the back-off model on the
largest set of data, it
outperforms the back-off model in almost all other cases. This clearly shows
the superior robustness of the
clustered models.
\begin{table}
\centering
\begin{tabular}{|c|cc|c|} \hline
Training & back-off & clustered & improvement (\%)\\ \hline
T1	&2240	&1750	&22\\
T2	&1130	&831	&27\\
T3	&621	&515	&17\\
T4	&373	&324	&13\\
T5	&254	&231	&9.1\\
T6	&190	&188	&1.1\\
T7	&159	&172	&-8.2\\ \hline
\end{tabular}
\caption{Perplexity results for (2000,2000) bigram clusters (VP)}
\label{tab:train_pp_vb}
\end{table}
\begin{table}
\centering
\begin{tabular}{|c|cc|c|} \hline
Training & back-off & clustered & improvement(\%)\\ \hline
T1	&3220	&2420	&25\\
T2	&1610	&1230	&24\\
T3	&878	&762	&13\\
T4	&537	&491	&8.6\\
T5	&369	&345	&6.5\\
T6	&265	&267	&-0.76\\
T7	&216	&240	&-11\\ \hline
\end{tabular}
\caption{Perplexity results for (2000,2000) bigram clusters (NVP)}
\label{tab:train_pp_nb}
\end{table}

Table \ref{tab:train_pp2} shows the results for a clustered tri-gram
\footnote{Only the 500,000 most frequent bigrams were clustered using
$G_{1}$.} with $7000$ and $1000$ clusters for $G_{1}$ and $G_{2}$ on VP data.
Because
these results were obtained on slightly different training and  testing texts,
the table also
contains the results of the clustered bi-gram on the same data. One
can see that the clustered trigram outperforms the
clustered bigram, at least with sufficient training data. But even with
only five million words of training data, the clustered trigram is only
slightly worse than the clustered bigram, showing again the
robustness of the clustered language models.
\begin{table}
\centering
\begin{tabular}{|c|cc|c|} \hline
Training & clustered bigram  & clustered trigram & improvement(\%) \\
 &  (3000,3000) & (7000,1000) & \\ \hline
5.2M	&191 &208 &-8.9\\
41M	&167 &151 &9.6\\ \hline
\end{tabular}
\caption{Results for bigram and trigram clusters (VP)}
\label{tab:train_pp2}
\end{table}

{}From all the results given here, one can see that the clustered
language models can still compete with unclustered models, even
when a large corpus, such as the Wall Street Journal corpus,
is being used.

\section{Conclusions}
\label{conclusions}

In this paper, an existing clustering algorithm is extended to deal with
higher order $N$-grams. Moreover, a heuristic version of the algorithm is
introduced, which leads to a very significant speed up (up to a factor
of 32),
with only a slight loss in performance (5\%). This makes it
possible to apply the resulting algorithm to the clustering of bigrams and
trigrams on the Wall Street Journal corpus. The results are shown to
be comparable to standard back-off bigram models.
Moreover, in the absence of many million words of training data, the clustered
model is more robust and clearly outperforms the non-clustered models.
This is an important point, because for many real world speech recognition
applications, the amount of training data available for a certain task or
domain is in general unlikely to exceed several million words. In those
cases, the clustered models seem like a good alternative to back-off models and
certainly one that deserves close investigation.

The main advantage
of the clustering models, its robustness in the face of little training
data, can also be seen from the results and in these situations, the
clustered algorithm is preferable to the standard back-off
models.

\section*{Appendix A: Deriving the Optimisation Function}
\label{derive}
In  this appendix, we will present the derivation of the optimisation function
for the extended clustering algorithm in detail. It is a generalisation of
\cite{Ueb94b},
where the derivation was given for $M=1$.

Let $G$ be a short hand to denote both classification functions $G_{1}$ and
$G_{2}$. Following the same approach as in section \ref{background}, one
can estimate the probabilities in equation \ref{eq:extending} using the
maximum likelihood estimator
\begin{eqnarray}
p_{G}(w|v_{M},...,v_{1}) & = & p(G_{2}(w)|G_{1}(v_{M},...,v_{1})) *
p(w|G_{2}(w))\\
 & = & \frac{N(g_{1}, g_{2})}{N(g_{1})} * \frac{N(g_{2}, y)}{N(g_{2})},
\end{eqnarray}
where $g_{1}=G_{1}(v_{M},...,v_{1})$, $g_{2}=G_{2}(w)$ and $N(x)$ denotes the
number
of times $x$ occurs in the data.
Given these probability estimates $p_{G}(w|v_{M},...,v_{1})$, the likelihood
$F_{ML}$
of the training data, e.g. the probability of the training data being generated
by our probability
estimates $p_{G}(w|v_{M},...,v_{1})$, measures how well the training data is
represented by the
estimates and can be used as optimisation criterion (\cite{Jel90}).
The likelihood of the
training data $F_{ML}$ is simply
\begin{eqnarray}
F_{ML} & = & \prod_{i=1}^{N} p_{G}(w_{i}|w_{i-M},...,w_{i-1})\\
 & = &  \prod_{i=1}^{N} \frac{N(G_{1}(w_{i-M},...,w_{i-1}),
G_{2}(w_{i}))}{N(G_{1}(w_{i-M},...,w_{i-1}))} * \frac{N(G_{2}(w_{i}),
w_{i})}{N(G_{2}(w_{i}))}.
\end{eqnarray}
Assuming that the classification is unique, e.g. that $G_{1}$ and $G_{2}$ are
functions, $N(G_{2}(w_{i}), w_{i})=N(w_{i})$ always holds
(because $w_{i}$ always occurs with the same class $G_{2}(w_{i})$). Since one
is trying to optimise $F_{ML}$ with respect to $G$, one
can remove any term that does not depend on $G$, because it will not influence
the optimisation. It is
thus equivalent to optimise
\begin{eqnarray}
F^{'}_{ML} & = & \prod_{i=1}^{N} \frac{N(G_{1}(w_{i-M},...,w_{i-1}),
G_{2}(w_{i}))}{N(G_{1}(w_{i-M},...,w_{i-1}))} * \frac{1}{N(G_{2}(w_{i}))}\\
 & =: & \prod_{i=1}^{N} f(w_{i-M},...,w_{i-1}, w_{i})  .
\end{eqnarray}
If, for two tuples $(w_{i-M},...,w_{i-1}, w_{i})$ and $(w_{j-M},...,w_{j-1},
w_{j})$, $G_{1}(w_{i-M},...,w_{i-1})=G_{1}(w_{j-M},...,w_{j-1})$
and $(G_{2}(w_{i})=G_{2}(w_{j})$ is true, then $f(w_{i-M},...,w_{i-1},
w_{i})=f(w_{j-M},...,w_{j-1}, w_{j})$ also holds.
One can thus regroup identical terms to obtain
\begin{equation}
 F^{'}_{ML}= \prod_{g_{1}, g_{2}} [ \frac{N(g_{1}, g_{2})}{N(g_{1})} *
\frac{1}{N(g_{2})} ] ^{N(g_{1}, g_{2})},
\end{equation}
where the product is over all possible pairs $(g_{1}, g_{2})$.
Because $N(g_{1})$ does not depend on $g_{2}$ and $N(g_{2})$ does not depend on
$g_{1}$, one can
simplify this again to
\begin{equation}
F^{'}_{ML}= \prod_{g_{1}, g_{2}} N(g_{1}, g_{2})^{N(g_{1}, g_{2})}
\prod_{g_{1}} \frac{1}{N(g_{1})}^{N(g_{1})}
\prod_{g_{2}} \frac{1}{N(g_{2})}^{N(g_{2})}.
\end{equation}
Taking the logarithm, one obtains the equivalent optimisation criterion
\begin{eqnarray}
 F^{''}_{ML} & = &  \sum_{g_{1}, g_{2}}  N(g_{1}, g_{2}) * log( N(g_{1}, g_{2}
)) -
\sum_{g_{1}} N(g_{1}) * log ( N(g_{1}))\\
 &  - &  \sum_{g_{2}} N(g_{2}) * log ( N(g_{2})) \nonumber .
\end{eqnarray}

$F^{''}_{ML}$ is the maximum likelihood optimisation criterion which could
be used to find a good classifications
$G$. However, the problem with this maximum likelihood criterion is the same as
in
section \ref{background}. In the following, a leaving-one-out
criterion is therefore developed.

Let $T_{i}$ denote the data without the pair $(w_{i-M},...,w_{i-1}, w_{i})$ and
$p_{G,T_{i}}(w|v_{M},...,v_{1})$ the probability estimates based on a given
classification $G$ and
training corpus $T_{i}$. Given a particular $T_{i}$, the probability of the
``held-out''
part $(w_{i-M},...,w_{i-1}, w_{i})$ is
$p_{G,T_{i}}(w_{i}|w_{i-M},...,w_{i-1})$. The probability of the complete
corpus,
where each pair is in turn considered the ``held-out'' part is the
leaving-one-out likelihood $L_{LO}$
\begin{equation}
 L_{LO}=\prod_{i=1}^{N} p_{G,T_{i}}(w_{i}|w_{i-M},...,w_{i-1}).
\label{eq:twob}
\end{equation}
In the following, we will derive an optimisation function $F_{LO}$ by
specifying how
$p_{G,T_{i}}(w_{i}|w_{i-M},...,w_{i-1})$ is estimated from frequency counts.
 One first rewrites $p_{G,T_{i}}(w_{i}|w_{i-M},...,w_{i-1})$  as usual (see
equation \ref{eq:extending}):
\begin{eqnarray}
 p_{G,T_{i}}(w|v_{M},...,v_{1}) & = &
P_{G,T_{i}}(G_{2}(w)|G_{1}(v_{M},...,v_{1}))*P_{G, T_{i}}(w|G_{2}(w))\\
 & = & \frac{p_{G,T_{i}}(g_{1}, g_{2})}{p_{G,T_{i}}(g_{1})} *
\frac{p_{G,T_{i}}(g_{2}, w)}{p_{G,T_{i}}(g_{2})},
\label{eq:estimb}
\end{eqnarray}
where $g_{1}=G_{1}(v_{M},...,v_{1})$ and $g_{2}=G_{2}(w)$.
Now we will specify how we estimate each term in equation \ref{eq:estimb}.

As before, $p_{G,T_{i}}$ can be dropped from the optimisation criterion and
relative frequencies can be used as estimators for the class unigrams:
\begin{eqnarray}
p_{G,T_{i}}(g_{1}) & = & \frac{N_{T_{i}}(g_{1})}{N_{T_{i}}} \label{eq:threeb}\\
p_{G,T_{i}}(g_{2}) & = & \frac{N_{T_{i}}(g_{2})}{N_{T_{i}}} \label{eq:fourb} .
\end{eqnarray}

In the case of the class bi-gram, one can again use the absolute discounting
method for smoothing. Let
$n_{0,T_{i}}$ be the number of unseen pairs $(g_{1}, g_{2})$ and $n_{+,T_{i}}$
the number of
seen pairs $(g_{1}, g_{2})$, leading to the following smoothed estimate
\begin{eqnarray}
\lefteqn{ p_{G,T_{i}}(g_{1}, g_{2})} \nonumber \\
 & = & \left\{ \begin{array}{ll}
\frac{N_{T_{i}}(g_{1}, g_{2}) - b}{N_{T_{i}}} & \mbox{if $N_{T_{i}}(g_{1},
g_{2})>0$}\\
\frac{n_{+, T_{i}}*b}{n_{0,T_{i}}*N_{T_{i}}} & \mbox{if $N_{T_{i}}(g_{1},
g_{2})=0$}\\
\end{array}
\right.
\end{eqnarray}
Again, the empirically determined constant value $b=0.75$  is used during
clustering.
The probability distribution $p_{G,T_{i}}(g_{1}, g_{2})$ will always be
evaluated on the
``held-out'' part $(w_{i-M},...,w_{i-1}, w_{i})$ and with
$g_{1,i}=G_{1}(w_{i-M},...,w_{i-1})$ and $g_{2,i}=G_{2}(w_{i})$ one obtains
\begin{eqnarray}
\lefteqn{ p_{G,T_{i}}(g_{1,i}, g_{2,i})} \nonumber \\
 & = & \left\{ \begin{array}{ll}
\frac{N_{T_{i}}(g_{1,i}, g_{2,i}) - b}{N_{T_{i}}} & \mbox{if
$N_{T_{i}}(g_{1,i}, g_{2,i})>0$}\\
\frac{n_{+, T_{i}}*b}{n_{0,T_{i}}*N_{T_{i}}} & \mbox{if $N_{T_{i}}(g_{1,i},
g_{2,i})=0$}
\label{eq:fiveb}
\end{array}
\right.
\end{eqnarray}

In order to facilitate future regrouping of terms, one again expresses the
counts
$N_{T_{i}}, N_{T_{i}}(g_{1})$ etc.
in terms of the counts of
the complete corpus $T$ as follows:
\begin{eqnarray}
N_{T_{i}} & = & N_{T} - 1 \label{eq:sixB} \\
N_{T_{i}}(g_{1}) & = & N_{T}(g_{1}) - 1 \\
N_{T_{i}}(g_{2}) & = & N_{T}(g_{2}) - 1 \\
N_{T_{i}}(g_{1,i}, g_{2,i}) & = &  N_{T}(g_{1,i}, g_{2,i})-1\\N_{T_{i}} & = &
N_{T} - 1 \\
n_{+, T_{i}} & = & \left\{
\begin{array}{ll}
n_{+, T} & \mbox{if $N_{T}(g_{1,i}, g_{2,i})>1$}\\
n_{+, T} - 1  & \mbox{if $N_{T}(g_{1,i}, g_{2,i})=1$}\\
\end{array} \right. \\
n_{0,T_{i}} & = &  \left\{
\begin{array}{ll}
n_{0,T} & \mbox{if $N_{T}(g_{1,i}, g_{2,i})>1$}\\
n_{0, T} - 1  & \mbox{if $N_{T}(g_{1,i}, g_{2,i})=1$}
\label{eq:sixbb}
\end{array} \right.
\end{eqnarray}
After dropping $p_{G,T_{i}}(w)$ and substituting the expressions back into
equation \ref{eq:twob}, one obtains:
\begin{eqnarray}
F'_{LO} & = & \prod_{i=1}^{N} \frac{p_{G,T_{i}}(g_{1,i},
g_{2,i})}{p_{G,T_{i}}(g_{1,i})} * \frac{1}{p_{G,T_{i}}(g_{2,i})}\\
& = & \prod_{g_{1}, g_{2}} ( p_{G,T_{i}}(g_{1}, g_{2}))^{N(g_{1}, g_{2})} *
\prod_{g_{1}} (\frac{1}{p_{G,T_{i}}(g_{1})})^{N(g_{1})} *
\prod_{g_{2}} (\frac{1}{p_{G,T_{i}}(g_{2})})^{N(g_{2})} .
\end{eqnarray}
One can now substitute equations \ref{eq:threeb}, \ref{eq:fourb} and
\ref{eq:fiveb}, using
the counts of the whole corpus of equations \ref{eq:sixB} to \ref{eq:sixbb} .
After having dropped
terms independent of $G$, one obtains
\begin{eqnarray}
F''_{LO} & = & \prod_{g_{1}, g_{2} : N(g_{1},g_{2})  > 1 } (N_{T}(g_{1}, g_{2})
-1 -b )^{N_{T}(g_{1}, g_{2})} *
\left( \frac{(n_{+,T}-1)*b}{(n_{0,T}+1)} \right)^{n_{1,T}}\\
 &  * & \prod_{g_{1}} \left( \frac{1}{(N_{T}(g_{1}-1))} \right)^{N_{T}(g_{1})}
* \prod_{g_{2}} \left( \frac{1}{(N_{T}(g_{2}-1))} \right)^{N_{T}(g_{2})}
\nonumber ,
\end{eqnarray}
where $n_{1,T}$ is the number of pairs $(g_{1}, g_{2})$ seen exactly once in
$T$
(e.g. the number of pairs that will be unseen when used as ``held-out'' part).
Taking the logarithm, one obtains the final optimisation criterion $F'''_{LO}$
\begin{eqnarray}
F'''_{LO} & = & \sum_{g_{1}, g_{2}:N_{T}(g_{1}, g_{2})>1} N_{T}(g_{1}, g_{2}) *
log ( N_{T}(g_{1}, g_{2}) - 1 - b )\\
 & + & n_{1,T} * log ( \frac{b*(n_{+, T}-1)}{(n_{0,T}+1)} ) \nonumber \\
 & - & \sum_{g_{1}} N_{T}(g_{1}) * log (  N_{T}(g_{1}) - 1 ) - \sum_{g_{2}}
N_{T}(g_{2}) * log (  N_{T}(g_{2}) - 1 )
\nonumber .
\end{eqnarray}


\end{document}